# Multiple Independent DE Optimizations to Tackle Uncertainty and Variability in Demand in Inventory Management

Sarit Maitra
*Business School*
*Alliance University*
Bengaluru, India
sarit.maitra@gmail.com

Sukanya Kundu
*Business School*
*Alliance University*
Bengaluru, India
sukanya.kundu@allience.edu.in

Vivek Mishra
*School of Applied mathematics*
*Alliance University*
Bengaluru, India
vivek.mishra@alliance.edu.in

*Abstract—* **To determine the effectiveness of metaheuristic Differential Evolution optimization strategy for inventory management (IM) in the context of stochastic demand, this empirical study undertakes a thorough investigation. The primary objective is to discern the most effective strategy for minimizing inventory costs within the context of uncertain demand patterns. Inventory costs refer to the expenses associated with holding and managing inventory within a business. The approach combines a continuous review of IM policies with a Monte Carlo Simulation (MCS). To find the optimal solution, the study focuses on meta-heuristic approaches and compares multiple algorithms. The outcomes reveal that the Differential Evolution (DE) algorithm outperforms its counterparts in optimizing IM. To fine-tune the parameters, the study employs the Latin Hypercube Sampling (LHS) statistical method. To determine the final solution, a method is employed in this study which combines the outcomes of multiple independent DE optimizations, each initiated with different random initial conditions. This approach introduces a novel and promising dimension to the field of inventory management, offering potential enhancements in performance and cost efficiency, especially in the presence of stochastic demand patterns.**

## I. INTRODUCTION

Inventory management (IM) is a critical aspect that is linked to business profitability in a modern organization. With increasing uncertainties and complexities, businesses need data-driven computational techniques to manage inventory. Real-world issues like stockouts, excess inventory, and revenue losses can be addressed using mathematical optimization approaches. In the past, several studies have made significant contributions to IM and highlighted the importance of sophisticated computational techniques to optimize inventory decisions to manage demand variations (e.g., [19]; [33]; [6] etc.). Building on their arguments, recent studies highlighted the growing complexities in IM driven by demand uncertainties, which have led to the development of computation-intensive simulation and optimization methods ([44] and [10]). Several recent studies have emphasized the relevance of optimization throughout the value chain (e.g., [24], [25], [11], etc.).

Despite several available works, the advancement of technology, globalization, and evolving customer expectations have made IM a complex task and an active research area. Researchers are constantly exploring innovative approaches and methodologies to handle this complexity effectively. Through this work, we address the questions of how to effectively manage inventory with stochastic demand, focusing on a continuous review policy approach, and how to optimize the total cost. Meta-heuristic optimization techniques such as Grey Wolf Optimizer (GWO), Whale Optimization Algorithm (WOA), Metaheuristics (MH) with Simulated Annealing (SA), Monte Carlo Simulation (MCS) with Bayesian Algorithm (BA), and Differential Evolution (DE) have been explored in this work. The findings reveal that DE is the most effective and simple heuristic optimization to deal with stochastic demands.

In the end, this work implements Adaptive DE by combining several DE variants and dynamically allotting computing resources based on their individual historical performance. The method helps to mitigate the risk of local optima and enhance the optimization process. The goal is to explore different regions of the parameter space to find a robust and reliable solution. The efficacy of the optimized policy may be sensitive to demand distribution. Therefore, this work performed a sensitivity analysis to assess the robustness of the policy under various scenarios.

The major contribution of this study is to experiment with a simulation-optimization model that can be applied with DE to select a nearly ideal IM policy under stochastic demand. The finding shows the proposed simulation optimization efficiently solves inventory policies by using the structure of the objective function rather than an exhaustive approach.

## II. PREVIOUS WORK

The importance of optimization is reflected in several studies (e.g., [25]; [24]; [11]; etc.). Optimization approaches offer a systematic method to further enhance inventory management ([31]; [28]). Studies (e.g., [33]; [34]; [26]) and subsequently industry reports suggest that IM costs can range a sizeable portion, which is approximately 20–40% of the total supply chain costs. It has been proposed that the effective simulation-optimization approach can bring a 16% reduction in costs by implementing the optimal policy. [12].

Several studies have discussed different optimization techniques ([14]; [4]; [42]). However, none of the works guaranteed an optimal solution if the original assumptions and considerations were violated [15]. On the same note, when using optimization techniques to solve IM problems, it is important to carefully consider the assumptions and constraints underlying the model [29]. These studies highlighted the importance of incorporating uncertainty and variability into the model because real-world inventory systems are often subject to such factors. Moreover, IM under uncertainty is challenging to solve due to the non-linearity of the model and several local optimum solutions ([8]; [17]). In recent times, metaheuristic algorithms have frequently been employed as powerful solutions for IM ([13]; [7]). Owing to their ability to effectively search for the solution space of complicated problems, meta-heuristic algorithms have received considerable attention in recent years (e.g., [1]; [9]; [41]; [7]; [34]; etc.). A recent study employed meta-heuristic algorithms for inventory optimization by presenting GWO and WOA as two novel solution approaches [30]. Though GWO was introduced to solve optimization problems [e.g., 23], some researchers criticized the fact that GWO mostly suffers from a lack of population diversity ([29]; [39]). To overcome this limitation, an improved version of GWO was presented [27]. Some authors emphasized the application of SA in the context of IM and the efficiency of SA in resolving the difficulties brought on by the unpredictability of demand and the requirement to optimize inventory policies under uncertainty (e.g., [40]; [21] etc.).

To overcome the limitations and improve algorithm efficiency, DE was introduced effectively for optimization [36]. It was compared with different optimization approaches; however, the DE method outperformed all other approaches in terms of the required number of function evaluations necessary to locate a global minimum of the test functions. Meta-heuristic approaches for inventory forecasting were also studied, which revealed the superior performance of DE even compared to CNN-LSTM [43]. The superiority of DE was supported by an exhaustive literature review, which revealed that 158 out of 192 papers were published between 2016 and 2021, showing that academics have improved DE to increase its effectiveness and efficiency in handling a variety of optimization challenges [2].

Moreover, the MCS method is commonly used to propagate the uncertainties of random inputs in the case of stochastic demand (e.g., [16]; [14]; [31]). This establishes that simulation is an integral part of IM during stochastic demands. MCS allows the incorporation of stochastic variability in demand patterns. The growing body of work in metaheuristic optimization indicates ongoing research efforts to improve the effectiveness of these techniques and their application in solving various optimization challenges in inventory management.

## III. METHODOLOGY

A three-stage approach employing a comprehensive methodology was adopted in this study to analyze and optimize the IM policy. Fig. 1 displays the methodological framework applied in this study, with shaded areas for various stages. First, the demand for products was collected over 365 days. The data were simulated to estimate the probability of experiencing various demand levels. These simulations allowed us to create multiple scenarios and observe the potential outcomes. The policies considered herein include continuous reviews and cross-docking. The performance of these policies was compared based on their ability to minimize total costs while ensuring an acceptable level of service.

$$Total_{cost} = Purchase_{cost} + Order_{cost} + Holding_{cost} + Stockout_{cost} \quad (1)$$

By considering all these costs, we aim to develop an inventory policy that minimizes costs and maximizes profits.

The results of the simulations are analyzed and interpreted to provide insights into the effectiveness of each policy. Once the optimal policy is identified, the next goal is to determine the optimal inventory levels that balance the $Total_{costs}$. With both the goals in place, in stage three (blue shaded area), the various optimization techniques (e.g., GWO, MH + SA, MCS, WO, MCS + BO, and DE) are employed.

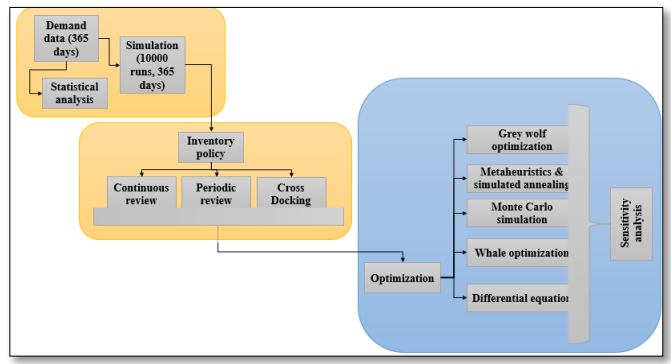

Fig. 1. Methodological framework (Source: Authors)

By employing the optimization technique and running simulations for a full year (365 days), we aim to fine-tune the inventory policy to minimize the $total_{costs}$ while considering the uncertainties in demand. In the final stage, sensitivity analysis is performed to identify different convergence rates, quality of solutions, and computational efficiency.

## IV. DATA ANALYSIS

The business case selected here examines the sale of four distinct products and considers the adoption of a suitable IM policy. The goal is to minimize the total cost associated with purchasing, ordering, and holding inventory by optimizing inventory levels. We use historical demand data to calculate the central tendency of the data. Table I displays the statistics related to the four products.

- Pr A, Pr B, Pr C, and Pr D are four distinct products.
- $Purchase_{costs}$ = cost of purchasing one unit of the item from the supplier.
- $Lead_{time}$ = time it takes for the supplier to deliver the item after placing an order.
- Size = size or quantity of each item.
- $Selling_{price}$ = price at which each item is sold to the customers.
- $Starting_{stock}$ = initial stock level of each item in the inventory.
- Mean = average demand for each item over a given period.
- $Std_{dev}$ = $Std_{dev}$ of demand for each item over a given period.
- $Order_{cost}$ = cost of placing an order with the supplier.
- $Holding_{cost}$ = cost of holding one unit of inventory for a given period.

- Probability = probability of a stock-out event occurring, i.e., the probability of demand exceeding the available inventory level.
- $Demand_{lead}$ = lead time demand for each item, i.e., the demand that is expected to occur during the lead time.

TABLE I. SUMMARY STATISTICS

|  | Pr A | Pr B | Pr C | Pr D |
|---|---|---|---|---|
| $Purchase_{cost}$ | € 12 | € 7 | € 6 | € 37 |
| $Lead_{time}$ | 9 | 6 | 15 | 12 |
| Size | 0.57 | 0.05 | 0.53 | 1.05 |
| $Selling_{price}$ | € 16.10 | € 8.60 | € 10.20 | € 68 |
| $Starting_{stock}$ | 2750 | 22500 | 5200 | 1400 |
| Mean | 103.50 | 648.55 | 201.68 | 150.06 |
| $Std_{dev}$ | 37.32 | 26.45 | 31.08 | 3.21 |
| $Order_{cost}$ | € 1000 | € 1200 | € 1000 | € 1200 |
| $Holding_{cost}$ | € 20 | € 20 | € 20 | € 20 |
| Probability | 0.76 | 1.00 | 0.70 | 0.23 |
| $Demand_{lead}$ | 705 | 3891 | 2266 | 785 |

Fig. 2 displays the KDE plots of the demand distribution of the products over 365 days. The shapes of the curves provide insight into the underlying stochastic distribution of the data. The isolated peaks in the curves show potential outliers in the demand data.

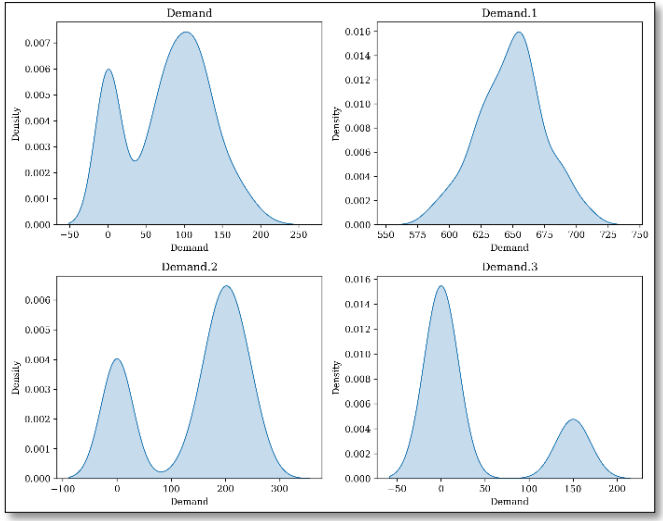

Fig. 2. KDE plots of demand distribution

*A. ABC analysis*

ABC analysis is performed to categorize items based on their initial IM values. The analysis follows the Pareto Principle for the annual consumption value of each product. Considering CV = consumption value, then,

$$CV_{annual} = Demand * Selling_{price} \quad (2)$$

$$CV_{cumulative} = \Sigma(CV_{annual}) \quad (3)$$

$$Cumulative\% = \Sigma(CV_{annual}/CV_{cumulative}) \quad (4)$$

- Product A: $CV_{annual} = Demand_{annual} * Selling_{price_{annual}} = 705 * €16.10 = €11,335.50$;
- Product B: $CV_{annual} = 3891 * €8.60 = €33,516.60$;
- Product C: $CV_{annual} = 2266 * €10.20 = €23,099.20$;
- Product D: $CV_{annual} = €53,380.00$.

$$CV_{cumulative} = CVA + CVB + CVC + CVD =$$
$$€11,335.50 + €33,516.60 + €23,099.20 +$$
$$€53,380.00 = €121,331.30 \quad (5)$$

- $Cumulative\%A = CVA / CV_{cumulative} = €11,335.50 / = €121,331.30 ≈ 0.0934$;
- $Cumulative\%B = CVB / CV_{cumulative} = €33,516.60 / = €121,331.30 ≈ 0.2763$;
- $Cumulative\%C = CVC / CV_{cumulative} = €23,099.20 / = €121,331.30 ≈ 0.1903$;
- $Cumulative\%D = CVD / CV_{cumulative} = €53,380.00 / = €121,331.30 ≈ 0.4399$.

ABC categories (A, B, and C) were assigned based on predefined cutoff points. Here, cutoff A was set at 0.8, and cutoff B was set at 0.95. Products with cumulative % below cutoff A were assigned category A, products with cumulative % between cutoff A and cutoff B were assigned category B, and products with cumulative % above cutoff B were assigned category C. Table II lists the metrics used in the analysis.

ABC analysis classifies products according to their consumption value, with Category A being the most critical products, Category B representing products of moderate importance, and Category C representing products of lower relevance.

*B. Latin Hypercube Sampling (LHS)*

The LHS is used to sample parameter combinations in a more evenly distributed manner. Table II lists the parameter values associated with the lowest average costs. The objective was to identify combinations that resulted in low total costs, indicating an efficient IM. The parameter space for calibration was taken as $reorder_{point}$ = {100, 200, 300}, $safety_{stock}$ = {50, 100, 50}, $lead_{time}$ = {0.8, 1.0, 1.2}, and $order_{quantity}$ = {0.8, 1.0, 1.2}. For experimentation and calibration, we treated $reorder_{point}$ and $safety_{stock}$ as separate entities. This helped for a comprehensive exploration of different inventory control strategies.

TABLE II. LATIN HYPERCUBE SAMPLING REPORT

|  | Pr A | Pr B | Pr C | Pr C | Lead time | Order Qty | Average cost |
|---|---|---|---|---|---|---|---|
| Reorder point | 753 | 6164 | 1425 | 383 | 0.8 | 0.8 | 77,540 |
| Safety stock | 377 | 3082 | 712 | 192 | | | |

The number of experiments was set at 27 because, based on orthogonal arrays, it represents the total number of unique combinations for the specified parameter space.

## V. SIMULATION & INVENTORY SYSTEMS

CR (continuous review) is more suitable for managing inventory with stochastic demand ([22]; [3]; [36]). An extensive literature review of over seven decades suggests that continuous policy is the most employed policy in stochastic inventory literature [28]. Taking a clue from their work, we examined both CR and CD in our empirical analysis. MCS was used to simulate and observe the cost and inventory levels over multiple simulation runs (10,000) for a period of 365 days.

Table III depicts the output; the reorder points and safety stocks for the analysis have been taken from Table II.

TABLE III. INVENTORY REVIEW SYSTEMS

| System definition | Average cost | Average inventory level |
|---|---|---|
| Continuous review | 515,262 | 21,251 |
| Cross docking | 515,268 | 21,253 |

Average cost and average inventory levels are calculated as:

$$average_{cost} = total_{cost} / (num_{simulations} * num_{periods})$$

$$average_{inventory\_level} = total_{inventory\_level} / (num_{simulations} * num_{periods})$$

The CD strategy involves minimizing the need for inventory storage by transferring products directly from the supplier to the customer. We implemented CD as additional logic within the IM simulation.

### A. Optimization

Multiple optimization algorithms (GWO, SA, MCS, MCS with BO, WO, and DE) were tested to get the optimal cost.

*1) Cost breakdown:*
The total cost is computed based on the following parameters:

- Purchase Cost is computed as $unit\ purchase_{cost} * max(order_{quantities}, 0)$.
- Order Cost is only applied when the order quantity is greater than zero and is computed as $unit\ order_{cost} * (order_{quantities} > 0)$.
- Holding Cost is computed as $unit\ holding_{cost} * max(inventory - demand, 0)$.
- Stockout Cost is inferred from the equation $unit\ holding_{cost} * max(demand - inventory, 0)$.

$$f(x) = \sum_{n}^{i=1}(PurchaseCost_i + OrderCost_i + HoldingCost_i + StockoutCost_i) \quad (6)$$

These costs can be subtracted from the revenue to give the corresponding profit for that one realization of the year. Eq. (7) formulates the annual profit, which is the future direction of this work:

$$SP_i \sum_{t=1}^{365} S_{i,t} - \left\{ \left(\frac{20V_i}{365}\right) \sum_{t=1}^{365} I_{i,t} + N_i C_{o,i} + \sum_{t=1}^{365} c_i P_{i,t} \right\} \quad (7)$$

Our goal is to minimize costs. Table IV presents a summary of all the algorithms tested on the given parameters and MCS to simulate the data for 365 days. We chose meta-heuristic techniques that are designed to tackle complex and non-linear optimization issues where typical optimization techniques struggle to find the global optimum.

TABLE IV. OPTIMAL POLICY FOR 365 DAYS

| Optimization | Stock | | | | Total Cost |
| --- | --- | --- | --- | --- | --- |
| | Pr A | Pr B | Pr C | Pr D | |
| GWO | 110 | 1836 | 0 | 21 | 17,391,348 |
| SA | 2,600 | 21,843 | 4,984 | 1,268 | 6,179,739 |
| MCS | 1,527 | 455 | 4,768 | 599 | 631,398 |
| WO | 1,070 | 10,865 | 3,787 | 150 | 504,939 |
| MCS with BO | 2,750 | 14,724 | 4,465 | 1,350 | 254,137 |
| DE (best1bin) | 1220 | 13204 | 3359 | 1317 | 250,774 |

Based on the $Total_{cost}$, the optimization method with the lowest total cost appears to be DE (best1bin), with the corresponding cost of 250,774. The future direction of our work is to check if DE can be further optimized.

### B. Multiple Independent DE Optimizations

In this approach, we performed optimization using multiple optimizers (five optimizers of DE with different random initial conditions) in parallel to determine the best parameters and cost. By using this approach, we aim to mitigate the impact of random variations in the MCS and increase the likelihood of finding a robust and optimal solution for IM. Considering, the below parameters:

- $purchase_{cost} = Pc_1, Pc_2, ..., Pc_n$, where, $Pc_i$ is the $purchase_{cost}$ of product $i$
- $lead_{time}$ = L$_1$, L$_2$, ..., L$_n$, where L$_i$ is the $lead_{time}$ of product i
- sizes = s$_1$, s$_2$, ..., s$_n$, where s$_i$ is the sizes of product i
- $selling_{price}$ = SP$_1$, SP$_2$, ..., SP$_n$, where SP$_i$ is the $selling_{price}$ of product i
- $starting_{stock}$ = ss$_1$, ss$_2$, ..., ss$_n$, where ss$_i$ is the initial inventory level of product i
- means = $\mu_1, \mu_2, ..., \mu_n$, where $\mu_i$ is the mean demand of product i;
- standard deviation = $\sigma_1, \sigma_2, ..., \sigma_n$, where $\sigma_i$ is the standard deviation of demand of product i
- order cost = $C_1, C_2, ..., C_n$, where $C_i$ is the order cost of product i
- $holding_{cost}$ = $V_1, V_2, ..., V_n$, where $V_i$ is the order cost of product $i$
- probabilities = $p_1, p_2, ..., p_n$, where $p_i$ is the probability of demand for product i
- demand lead = $D_1, D_2, ..., D_n$, where $D_i$ is the demand lead time of product i.

- Parameters space of optimization: $bounds = [(0, ss_1), (0, ss_2) \ldots (0, ss_n)]$, where each tuple represents the lower and upper bounds of inventory levels of the respective products.

MCS and objective function (x), where $x = x_1, x_2, x_3, \ldots, x_n$ representing inventory levels.

$$reorder\ levels = [means1 * L_1 + \sqrt{L_1} * \sigma1, \ldots, meansn * \sqrt{L_n} * \sigma n]$$

$$order_{quantity} = [\max(reorder_{level_1} - x_1, 0), \ldots, \max(reorder_{level_n} - x_n, 0)]$$

$total_{cost}$ = for each day and product: if $x_i < reorder_{level}$, $order_{quantity} = order_{quantity_i}$ (this checks if the current inventory level is below the reorder level). If it is, the order quantity is set to the predetermined value for that product ($order_{quantity_i}$), indicating that an order should be placed to replenish the inventory.

- $increaseInventory\ (xi) = xi + orderQuantity$
- $increaseTotalCost = totalCost +$
- $decreaseInventory\ (x_i) = x_i - daily_{demand}$, if $x_i < 0$, $x_i = 0$ and $totalCost = totalCost + holdingCost/2$
- if $(d + 1)\%\ lead_{times_i} = 0$, decrease inventory again: $xi = xi - daily_{demand_i}$ and $totalCost = totalCost + holding_{cost_i} * x_i$

$$mean_{cost} = \frac{1}{num_{samples}} \sum_{s=1}^{num_{samples}} total_{cost_s} \quad (8)$$

$$MultipleDE_{optimization} = [result_1, result_2, \ldots, result_{num_{ensemble}}] \quad (9)$$

where, each result represents the optimization result of one DE member. Table V reports the output.

TABLE V. MULTIPLE INDEPENDENT DE OPTIMIZATIONS

| Optimization | Stock | | | | Total Cost |
| --- | --- | --- | --- | --- | --- |
| | Pr A | Pr B | Pr C | Pr D | |
| Multiple Independent DE | 2567 | 9063 | 4277 | 1322 | 249,128 |

The total cost has been marginally reduced from 250,774 (Table IV) to 249,128 (Table V). The stocks are optimized from 19,100 (1220, 13204, 3359, 1317) to 17,229 (2,567, 9,063, 4,277, 1,322). The mutation rate and crossover rate are adaptively adjusted during the process based on iteration success or failure, ensuring a balance between exploration and exploitation during optimization.

C. Sensitivity analysis

Sensitivity analysis is employed on this to ensure the robustness of the multiple independent DE optimizations model under different scenarios. The analysis was performed on the population size parameters of the DE algorithm. The goal was to evaluate the effects of different population sizes on the optimization results. Different values of population size, for example, 10, 20, 50, and 100, were tested to observe how they affected the optimization results. By exploring different population sizes, we have assessed their impact on convergence behavior and the quality of the obtained solutions.

TABLE VI. SENSITIVITY ANALYSIS

| Analysis | Stock | | | | Total Cost |
| --- | --- | --- | --- | --- | --- |
| | Pr A | Pr B | Pr C | Pr D | |
| Population size 10 | 2,271 | 4,736 | 4,146 | 1,321 | 251,238 |
| Population size 20 | 1,901 | 20,134 | 3,355 | 1,325 | 249,780 |
| Population size 50 | 1,525 | 12,753 | 4,992 | 1,326 | 246,251 |
| Population size 100 | 1,552 | 9,667 | 3,695 | 1,326 | 246,745 |

By varying the population size in the DE, we can observe how it affects the optimization results. In this case, the observed differences in total cost are small, indicating that the model's performance is stable and not heavily influenced by changes in population size.

D. Critical findings

This study emphasizes the importance of optimization in IM, specifically in the context of stochastic demand and supply disruptions. DE is a successful method for establishing near-optimal inventory policies when combined with best/1/bin mutation strategy, LHS, and multiple independent DE optimizations. Sensitivity analysis with varying population sizes confirmed the stability of the optimization model.

VI. CONCLUSION

This study highlighted the significance of optimization techniques, particularly DE and multiple independent DE optimizations, in achieving cost-effective and robust inventory management strategies in the face of uncertain demand and supply disruptions. Empirical analysis was conducted using 365-day demand data and reported the optimal policy, along with cost comparisons. The study also discussed the use of LHS for efficient parameter sampling. ABC analysis was applied to categorize items and assign Pareto classes to products. The optimal policy and inventory levels were determined through simulations and optimization techniques. A sensitivity analysis assessed convergence rate, solution quality, and computational efficiency. This comprehensive approach contributes to IM by improving efficiency and cost-effectiveness while addressing demand uncertainties.